\documentclass[conference,9pt]{IEEEtran}
\IEEEoverridecommandlockouts
\usepackage{cite}
\usepackage{amsmath,amssymb,amsfonts}
\usepackage{algorithmic}
\usepackage{graphicx}
\usepackage{textcomp}
\usepackage{xcolor}
\usepackage{tabularray}
\usepackage[normalem]{ulem}
\usepackage{booktabs}
\usepackage{multirow}
\def\BibTeX{{\rm B\kern-.05em{\sc i\kern-.025em b}\kern-.08em
    T\kern-.1667em\lower.7ex\hbox{E}\kern-.125emX}}
\begin{document}

\title{Focusing Image Generation to Mitigate Spurious Correlations\\
\thanks{*Corresponding author: Zhiqiang Liu (tjubeisong@tju.edu.cn).}
}

\author{\IEEEauthorblockN{\textit{ Xuewei Li$^{1,2,3,4}$, Zhenzhen Nie$^{1,2,3}$, Mei Yu$^{1,2,3,4}$,  Zijian Zhang$^{1,2,3}$, Jie Gao$^{1,2,3}$,  Tianyi Xu$^{1,2,3}$, Zhiqiang Liu$^{\ast,1,2,3}$}
}

\IEEEauthorblockA{
$^{1\;}$College of Intelligence and Computing, Tianjin University, Tianjin, 300350, China.\\
$^{2\;}$Tianjin Key Laboratory of Cognitive Computing and Application, Tianjin, 300350, China.\\
$^{3\;}$Tianjin Key Laboratory of Advanced Networking, Tianjin, 300350, China.\\
$^{4\;}$School of Feature Technology, Tianjin University, Tianjin, 300350, China.}}

\maketitle

\begin{abstract}
Instance features in images exhibit spurious correlations with background features, affecting the training process of deep neural classifiers. This leads to insufficient attention to instance features by the classifier, resulting in erroneous classification outcomes. In this paper, we propose a data augmentation method called Spurious Correlations Guided Synthesis (SCGS) that mitigates spurious correlations through image generation model. This approach does not require expensive spurious attribute (group) labels for the training data and can be widely applied to other debiasing methods. Specifically, SCGS first identifies the incorrect attention regions of a pre-trained classifier on the training images, and then uses an image generation model to generate new training data based on these incorrect attended regions. SCGS increases the diversity and scale of the dataset to reduce the impact of spurious correlations on classifiers. Changes in the classifier’s attention regions and experimental results on three different domain datasets demonstrate that this method is effective in reducing the classifier’s reliance on spurious correlations.
\end{abstract}


\section{Introduction}
As one of the core technologies in the field of artificial intelligence, neural networks have achieved remarkable progress in image classification and other domains. However, despite their notable performance on many tasks, these classifiers still face the problem of spurious correlations which are quite common in real-world datasets\cite{paper1,paper2}. Spurious correlations, namely “correlations that do not imply causation” in statistics, are brittle associations between certain attributes of inputs and target variable\cite{paper25,paper27}. These attributes are referred to as spurious attributes. Spurious correlations may lead the model to perform well on the training data but poorly on new, unbiased data. For example, in a binary classification task between ``cat” and ``dog”, the dataset may contain many images of ``cat” on ``bed” and ``dog” on ``bench”, leading the neural networks to rely more on the presence of ``bed” or ``bench” \cite{paper10} for classification rather than the actual features of the ``cat” or ``dog”.

\begin{figure}
	\centerline{\includegraphics[scale=0.35]{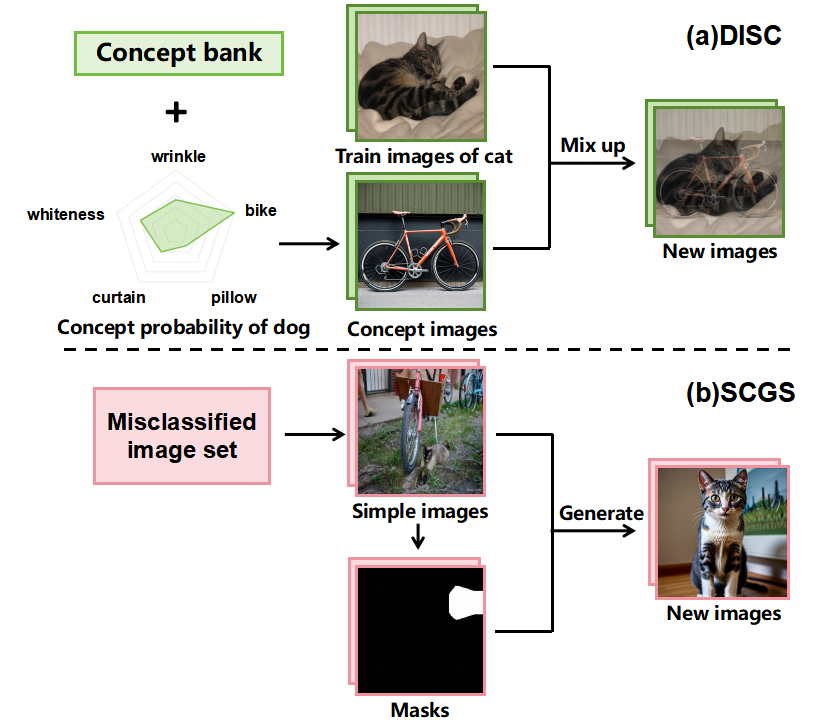}}
	\caption{(a) Existing DISC\cite{paper9} method generates new images based on a concept bank and concept probabilities. (b) SCGS generates new images by using labels, masks, and misclassified images from training set as inputs for an existing image generation model\cite{paper6}.}
	\label{fig1}
\end{figure}

In recent years, various solutions have been proposed to mitigate the negative impacts of spurious correlations. Methods like Group DRO \cite{paper4,paper7} utilizes regularization techniques to optimize models, ensuring good performance even when group distributions shift. Zhang Y et al. proposed a method of generating adversarial samples to enhance the training data\cite{paper10,paper12}. Additionally, DFR\cite{paper13} enhances model robustness by retraining the final layer of a model that previously trained using Empirical Risk Minimization (ERM). As shown in Fig. 1, DISC\cite{paper9} iteratively identifies unstable concepts (spurious attributes) across different environments by constructing a concept bank. Subsequently, it mixes concepts with training data to intervene in the data distribution. However, these methods require the addition of group labels to the training set or an additional concept bank, both of which are often expensive to get. Moreover, the reliance on group labels and concept banks confine the identification of spurious attributes to those that are given, potentially limiting the discovery of other spurious attributes. Liu E Z and others propose methods that do not require adding group labels to the training set, such as re-weighting biased data by training a biased network to identify the bias in the data (e.g. JTT)\cite{paper5,paper8}. However, due to the small number of images without spurious correlations, the model is prone to over-fitting and fails to learn instance features\cite{paper28}. Mitigating the impact of potential spurious correlations from the data while learning robust classifiers remains a challenging task.

In this paper, we propose a simple yet effective image classification framework termed Spurious Correlations Guided Synthesis (SCGS), which generates new data to augment the training dataset without relying on group labels or a concept bank. Due to the standard ERM model often exhibiting spurious correlations when trained on biased datasets (with lower accuracy in the worst-group), SCGS focuses on analyzing misclassified images from the ERM model. Specifically, SCGS is divided into three stages: we first identify the misclassified training data under the standard ERM model and perform clustering and sampling. Then, we obtain masks through class activation maps\cite{paper29} of the sampled images. Finally, we generate new images based on the masks, sampled images, and image true labels. These new images are then added into the original dataset, allowing for further model training. Experimental results on three datasets with spurious correlations: MetaShift\cite{paper14}, Waterbirds\cite{paper7}, and CelebA\cite{paper16}, indicate that SCGS performs well on all three datasets. When applied to JTT, SCGS shows a 4.9\% improvement over other methods on the MetaShift dataset. Additionally, SCGS also achieves competitive results on the Waterbirds and CelebA datasets. Our contribution can be summarized as follows:

\begin{itemize}
\item We develop SCGS, a novel and effective framework to guide the model to learn instance features, which extracts spurious attributes through clustering, effectively reducing the impact of spurious correlations on model training.
\item We propose a minority image generation method that generates new images directly from existing images, without the need for group labels or a concept bank.
\end{itemize}

\section{Method}
In this section, we describe our method in detail. We firstly introduce the problem setup in Subsection A and provide an overview in Subsection B. Then, we discuss the clustering and sampling of misclassified images in Subsection C. In Subsection D, we describe the generation of class activation maps and masks. Finally, we introduce the generation of new images and the mitigation of spurious correlation effects in Subsection E.

\subsection{Problem Setup}
We consider a classification task with a training dataset $D_{tr}=\left \{ \left ( x_{i},y_{i} \right )  \right \} _{i=1}^{N}$, where $x_{i}\in X$ is the training image and $y_{i}\in Y$ is the image label. During training, since an attribute $a_{1}\in A$ co-occurs with a label $y_{1}\in Y$, the classifier mistakenly believes that the attribute is associated with the label. This connection is a spurious correlation ($a_{1}\sim y_{1}$), as there is no direct causal relationship between them. Due to the presence of spurious correlations, the classifier incorrectly classifies the test image $\left ( x_{2},y_{2} \right )$ in which attribute $a_{1}$ appears as $y_{1}$. The classifier no longer focuses on the instance features, but rather focus more on attribute $A$, which is used as the basis for its classification decisions.

The classification model is defined as $f_{\theta }:X\to Y$, where $\theta$ is a set that includes all the parameters of the model. For the loss function $\ell \left ( x,y;\theta  \right )$, the objective of Empirical Risk Minimization (ERM) is to find a set of parameters $\theta ^{\ast }$ that minimizes the empirical loss for the model: 

\begin{equation}
\theta ^{\ast } = \mathop{\arg\min}\limits_{\theta} \mathbb{E}_{\left ( x,y \right ) \sim D_{tr}}\left [ \ell \left ( x,y;\theta  \right ) \right ] 
\end{equation}

If the training dataset is biased, a model trained with ERM may misclassify on the unbiased test set due to the presence of spurious correlations. In this paper, SCGS method aims to obtain a classifier $f_{\theta }^{'}$ that focuses more on the instance features rather than the spurious attributes, and performing well on both the training set and test set. Notably, this task does not require group labels from the training dataset.

\subsection{Overview}
As shown in Fig. 2, SCGS first trains the ERM to obtain a classifier $f_{\theta }$, obtains the misclassified image set $D_{mis}$ in the training set. Then, K-means clustering and Gaussian sampling are applied to obtain the sampled set $D_{sam}$. Next, Grad-CAM++\cite{paper19} is used to generate class activation maps and corresponding masks for the sampled set. Finally, based on the original images, corresponding masks, and image labels, Stable Diffusion\cite{paper6} is employed to generate new images to balance the dataset for updating the classifier parameters.

\begin{figure*}
	\centerline{\includegraphics[scale=0.35]{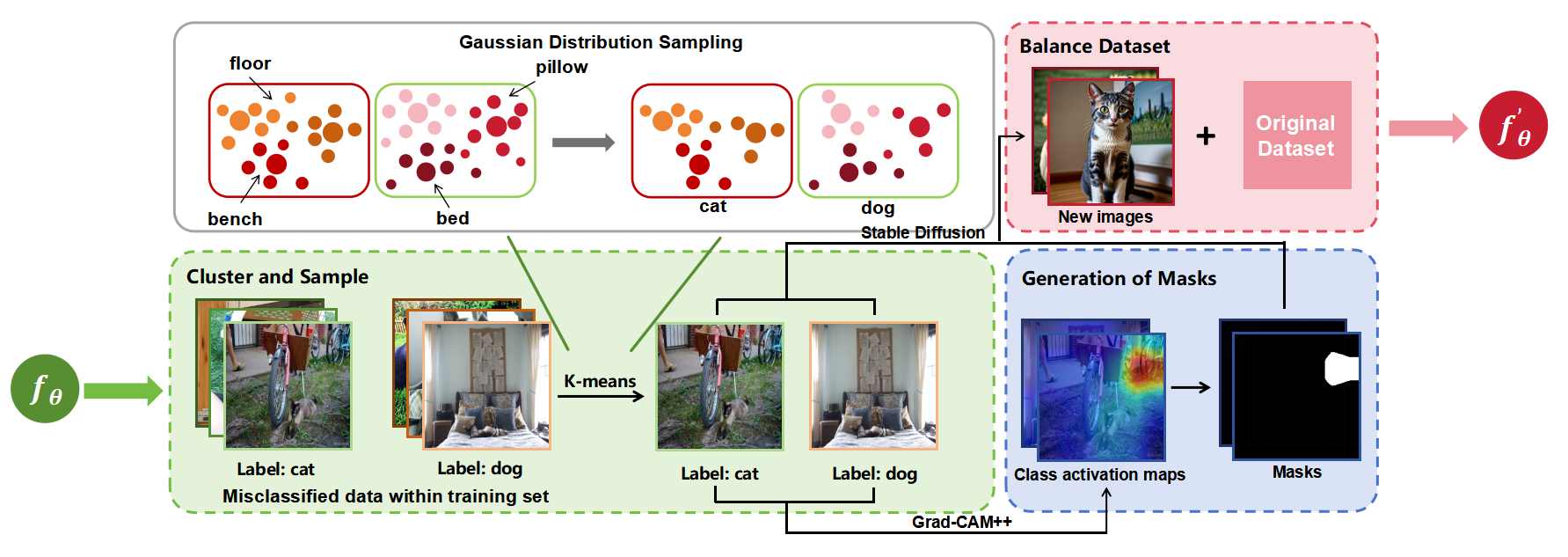}}
	\caption{Overview of our SCGS.}
	\label{fig3}
\end{figure*}

\subsection{Cluster and Sample}
Based on the observation, the simpler ERM model is able to fit the data with easily recognizable spurious correlations. For example, in the MetaShift dataset, ``cat” frequently appear with ``bed” and ``sofa”, while ``dog” often appear with ``chair”. But for the data that do not show spurious correlations, such as ``dog” on ``bed”, achieving effective fitting is a significant challenge for ERM models. Therefore, SCGS first trains the ERM model and employs it to identify misclassified images in the training dataset. Then, obtaining the labels and prediction results of the misclassified image set $D_{mis}$. The set of misclassified image set for each class can be represented as:

\begin{equation}
D_{mis}^{i} = \left \{ \left ( x,y_{i} \right )\mid f_{\theta}\left ( x \right ) \ne y_{i} \right \},i=1,...,N  
\end{equation}

Where $i$ is the class, and $N$ is the number of class. For images within the same class, the spurious correlations that lead to misclassification by the ERM model are maybe different. For example, a cat might be mistakenly classified as a dog, because the cat’s image contains an attribute that has a spurious correlation with dog. However, there are various attributes have spurious correlations with dog, such as ``tree” or ``chair”. Moreover, for larger datasets, the number of misclassified images is larger, which increases the cost of generating images. Therefore, we cluster the images of each class based on their feature vectors, aiming to organize them according to the types of spurious correlations. Then, we sample from each cluster following the principles of Gaussian distribution to obtain a reasonably sized set of images $D_{sam}$.

Since the k-means clustering algorithm\cite{paper17} has explicit cluster centers, SCGS utilizes the k-means algorithm for clustering. For each cluster $k$ in class $i$, the feature vectors within it are defined as a set $S_{k}$, and the mean of these feature vectors is defined as the cluster centroid $\mu _{k}$. By calculating the covariance matrix $\Sigma _{k}$ for each cluster $k$ and the gaussian probability density function value $P\left ( s\mid \mu _{k},\Sigma _{k} \right )$ for each feature vector $s\in S_{k}$, a certain number of images can be sampled from each cluster. As shown in the 'Gaussian Distribution Sampling' section of Fig. 2. 

For each image $x$, its Gaussian probability density function value is shown below, which serves as the probability of each image being extracted:

\begin{equation}
\begin{array}{c}   P\left ( s\mid \mu _{k},\Sigma _{k} \right ) = \frac{exp\left ( - \frac{1}{2} \left ( s-\mu _{k} \right )^{T} \Sigma _{k}^{-1}\left ( s-\mu _{k} \right )  \right ) }{\sqrt{\left ( 2\pi  \right )^{D}\left | \Sigma _{k} \right |  } }\\    \text{where }\Sigma _{k} = \frac{\Sigma_{i =1}^{\left | S_{k} \right | }\left ( s_{i}-\mu _{k} \right )\left ( s_{i}-\mu _{k} \right )^{T}  }{\left | S_{k} \right |-1 } \end{array}
\end{equation}

Here, $\left | S_{k} \right | $ is the number of feature vectors in cluster $k$, $s$ is the feature vector of image $x$, $D$ is the dimension of the feature vectors, and $\left | \Sigma _{k} \right |$  is the determinant of covariance matrix.

Finally, we obtain the sample image sets for each class. The sample image set for class $i$ can be represented as follows, where $K$ is the number of clusters:
\begin{equation}
D_{sam}^{i}=\cup _{k=1}^{K}D_{sam}^{i\left ( k \right ) }
\end{equation}

\subsection{Generation of Masks}
Gradient-weighted Class Activation Mapping(Grad-CAM)\cite{paper18} is a visualization method used to interpret the decision-making process of deep learning models, particularly convolutional neural networks (CNNs). It generates class activation maps from the gradient of the last convolutional layer of a CNN, helping us understand which regions of the image are more important for the network's predictions. Grad-CAM++\cite{paper19} is an improved version of the Grad-CAM method. It considers that each element on the gradient map contributes differently, so an additional weight is applied to the elements on the gradient map to achieve more accurate localization. The process of obtaining the class activation maps can be represented as:

\begin{equation}
\begin{array}{c} H\left ( x \right ) =Grad-CAM\left ( x,f_{\theta }\left ( x \right ), \theta \right ) \\    \text{where }f_{\theta }\left ( x \right )\ne y,\left ( x,y \right )\in D_{sam} \end{array}
\end{equation}

For the sample image $x$, the prediction result from ERM model differs from the true label. Therefore, the Grad-CAM++ method is employed to generate a class activation map based on this incorrect prediction $f_{\theta }\left ( x \right )$, ERM model parameters $\theta$ and image $x$, visually explaining the reason for the misclassification of the ERM model. By setting a threshold, we divide the class activation map into high-contribution regions and low-contribution regions, considering that the high-contribution regions have spurious correlations with incorrect prediction results. To retain the high-contribution regions, we mark these regions as 1 in the mask, and mark the other regions as 0.

\subsection{Balance Dataset}
An important reason for the emergence of spurious correlations during model training is that a spurious attribute appears frequently in images of one class, while rarely appearing in images of other classes. Therefore, we associate the spurious attribute with other classes, generate more new images that are under-represented in the training data. Then, we add these new images to the original dataset to balance the original data, guiding the model to focus more on the object itself. For example, since ``cat" frequently appears with ``bed", the model incorrectly classifies ``dog" as ``cat" when ``dog" and ``bed" appear together. To address this, more images of ``dog" with ``bed" are generated and added to the training dataset to influence the model's prediction.

SCGS uses masks obtained from Subsection D, along with the corresponding original images and their labels (as prompts) to generate new images. This work is accomplished by Stable Diffusion\cite{paper6}, which possesses powerful image generation capabilities. Stable Diffusion consists of three components: a text encoder CLIP\cite{paper22}], a diffusion model (U-net\cite{paper23} + Scheduler), and a Variational Autoencoder (VAE). In this paper, masks indicate which areas of the image should be retained when generating a new image, while the labels define the content that should be generated in the new image. It should be noted that the white areas of the mask are considered the regions to be preserved, while the black areas are the regions that need to be generated or changed.

\begin{table*}[h]
	\centering
	\caption{\textnormal{Overall experimental results. The best results are shown in \textbf{bold} and the second-best results are \uline{underlined}. \\ The "Train group info" column indicates whether group labels is used in the training dataset.}}
	\label{tab:results}
    \renewcommand\arraystretch{1.2}
	\begin{tabular}{lccccccc}
\toprule
 & \textbf{Train} & \multicolumn{2}{c}{Metashift} & \multicolumn{2}{c}{Waterbirds} & \multicolumn{2}{c}{CelebA} \\
 \cmidrule(lr){3-4} \cmidrule(lr){5-6} \cmidrule(lr){7-8}
 & \textbf{group info} & Avg Acc. & Worst-group Acc. & Avg Acc. & Worst-group Acc. & Avg Acc. & Worst-group Acc. \\
\midrule
		Group DRO\cite{paper7} & Yes & $73.6\pm2.1$ & $66.0\pm3.8$ & $91.8\pm0.3$ & $90.6\pm1.1$ & $92.9\pm0.3$ & $86.3\pm1.1$ \\
		LISA\cite{paper11} & Yes & $70.0\pm0.7$ & $59.8\pm2.3$ & $91.8\pm0.3$ & $88.5\pm0.8$ & $92.4\pm0.4$ & $89.3\pm1.1$ \\
		DISC\cite{paper9} & Yes & $75.5\pm1.1$ & $73.5\pm1.4$ & $93.8\pm0.7$ & $88.7\pm0.4$ & - & - \\
		
  \bottomrule
        LfF\cite{paper24} & No & $\mathbf{80.8\pm0.1}$ & $62.7\pm2.3$ & \uline{$91.2$} & $78.0$ & $85.1$ & $77.2$ \\
		MaskTune\cite{paper21} & No & - & - & $\mathbf{93.0\pm0.7}$ & \uline{$86.4\pm1.9$} & $91.3\pm0.1$ & $78.0\pm1.2$ \\
		ERM & No & $73.6\pm1.1$ & $62.2\pm2.2$ & $87.1\pm0.8$ & $68.8\pm0.6$ & $\mathbf{94.6\pm0.2}$ & $45.6\pm1.4$ \\
		\textbf{SCGS} & No & \uline{$79.4\pm1.4$} & \uline{$75.3\pm1.6$} & $90.6\pm0.3$ & $77.9\pm1.2$ & \uline{$91.5\pm0.8$} & $69.6\pm0.4$ \\
		JTT\cite{paper8} & No & $75.6\pm0.6$ & $66.2\pm1.3$ & $89.9\pm0.5$ & $86.4\pm1.4$ & $88.1\pm0.3$ & \uline{$81.5\pm1.7$} \\
		\textbf{JTT+SCGS} & No & $79.2\pm0.9$ & $\mathbf{78.4\pm1.6}$ & $90.9\pm0.3$ & $\mathbf{88.3\pm0.9}$ & $89.8\pm0.4$ & $\mathbf{82.5\pm1.3}$ \\
  \bottomrule
	\end{tabular}
\end{table*}

\begin{table*}[h]
\centering
\caption{\textnormal{Performance of three different image generation methods.}}
\label{tab:comparison}
\renewcommand\arraystretch{1.2}
\begin{tabular}{lccccccc}
\toprule
 & \multicolumn{2}{c}{Img2img} & \multicolumn{2}{c}{Grad-Cam} & \multicolumn{2}{c}{Grad-Cam++} \\
\cmidrule(lr){2-3} \cmidrule(lr){4-5} \cmidrule(lr){6-7}
 & Avg Acc. & Worst-group Acc. & Avg Acc. & Worst-group Acc. & Avg Acc. & Worst-group Acc. \\
\midrule
Metashift & $77.0\pm0.8$ & $68.8\pm2.2$ & $78.3\pm0.7$ & $71.0\pm1.8$ & $79.4\pm1.4$ & $75.3\pm1.6$ \\
Waterbirds & $87.6\pm0.7$ & $73.5\pm0.5$ & $88.9\pm0.4$ & $75.5\pm0.9$ & $90.6\pm0.3$ & $77.9\pm1.2$ \\
CelebA & $93.9\pm0.4$ & $64.5\pm1.1$ & $92.3\pm0.6$ & $65.1\pm0.7$ & $91.5\pm0.8$ & $69.6\pm0.4$ \\
\bottomrule
\end{tabular}
\end{table*}

\section{Experiment}

\subsection{Experimental Setup}
\textbf{Datasets.} We evaluate the SCGS method on three datasets. (a) MetaShift dataset consists of 1664 images of various types of cats and dogs, where ``cat" and ``dog" have spurious correlations with certain attributes (e.g. bed, chair). (b) Waterbirds dataset contains a total of 11,788 images of two types: waterbird and landbird, and the number of ``landbird" with a ``water" background and ``waterbird" with a ``land" background is relatively small.(c) CelebA dataset consists of 202,599 images of celebrity faces, and the task is to classify the hair color of the people into ``blond" and ``not blond". Among the dataset, the proportion of men with blond hair is relatively small.

\textbf{Model Training.} In this paper, we use a pre-trained ResNet-50\cite{paper26} model for training. During clustering, the MetaShift dataset is divided into 4 clusters per class, while Waterbirds and CelebA are set to 2 clusters per class. The sample size is set to 20\% of the misclassified images in each class. In mask generation, red and orange regions are considered as high-contribution regions. The newly generated images per class for the MetaShift and Waterbirds datasets are set to 40\% of the images in other classes, while  for the CelebA dataset, it is set to 20\%. We apply the SCGS method to ERM and JTT[8] to verify its effectiveness, with batch sizes set to 16 for MetaShift, 64 for Waterbirds, and 128 for CelebA. The settings for JTT follow those of the original author.

\subsection{Experimental Result}
We compare SCGS with standard ERM and recent methods. TABLE I presents the detailed results. In these experiments, both LISA and Group DRO models utilize group labels during training, and DISC applies group labels on the Waterbirds dataset. The results for LISA and Group DRO on Metashift and Waterbirds are based on digits provided in the DISC. All methods use the same data splits and evaluation criteria, including average accuracy and worst-group accuracy (each group is defined by instances and spurious attributes, such as waterbird-water, waterbird-land). For all datasets, the worst-group accuracy of the JTT+SCGS model is closer to the results of methods using group labels, and it even surpasses them on the Metashift dataset, with a significant improvement of 4.9\% in worst-group accuracy. Compared to standard ERM, SCGS significantly improves worst-group accuracy (by 13.1\%, 9.1\%, and 9.7\% for each dataset), while also enhancing average accuracy on Metashift and Waterbirds (by 5.8\% and 3.5\%). For the JTT+SCGS model, compared to the JTT model, there are notable improvements in worst-group accuracy across datasets (by 12.2\%, 1.9\%, and 1\%). These results clearly demonstrate that SCGS has excellent debiasing capabilities. Furthermore, the improvement of SCGS on the CelebA dataset is less evident, which may be due to the limited impact of additional images on the overall distribution when the dataset is already sufficiently large.

\begin{figure}
	\centerline{\includegraphics[scale=0.24]{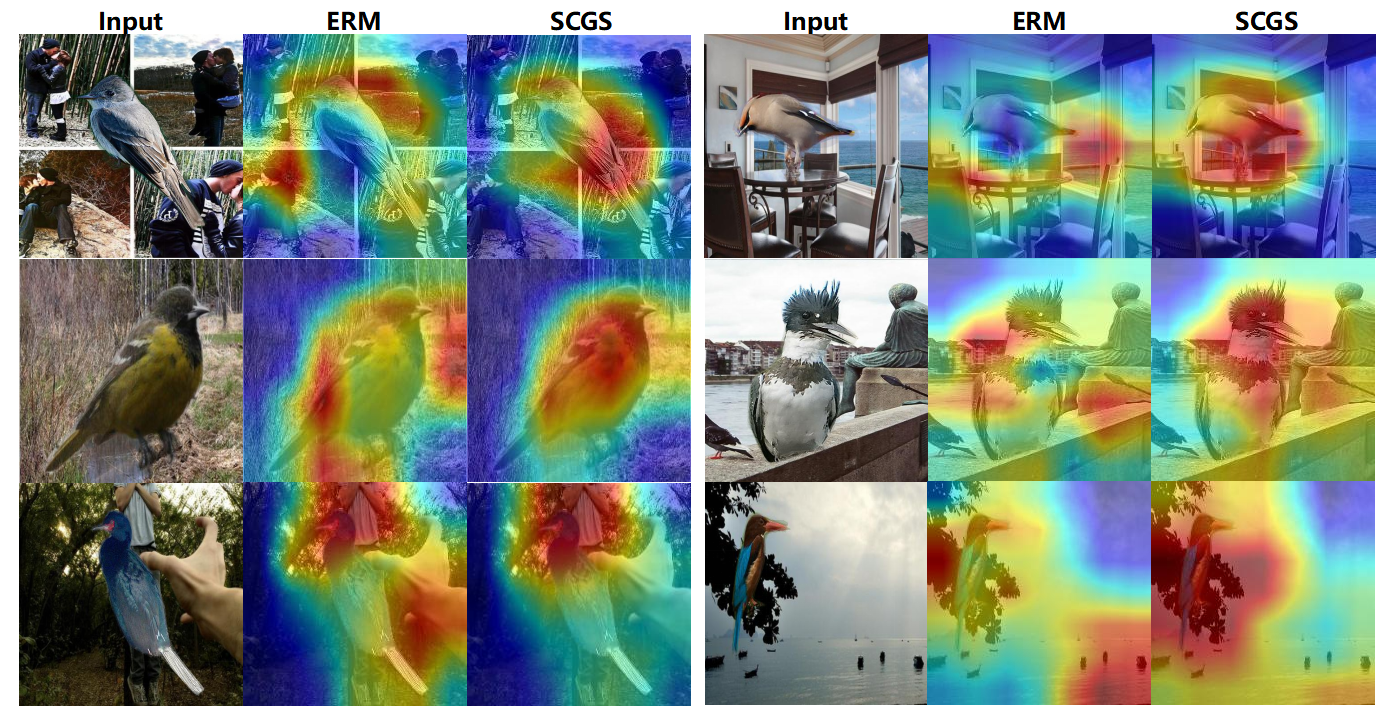}}
	\caption{Grad-CAM++ comparison of ERM and SCGS.}
	\label{fig4}
\end{figure}

To evaluate the impact of different images genetation methods on the classification capabilities of models, we generate new images using three methods: directly generating new images from sampled images using stable diffusion(img2img), generating new images using stable diffusion(inpaint upload) with class activation maps from Grad-CAM and Grad-CAM++ as masks. Experiments on three datasets in TABLE II show that generating new images using class activation maps produced by Grad-CAM++ achieves the best results, while generating images directly from sampled images yields the poorer results. This clearly demonstrates the effectiveness of using class activation maps for masks generation.

We also observe changes in class activation maps on images from the test set to gain a more intuitive understanding of SCGS's debiasing capability. As shown in Fig. 3, it can be seen that ERM focuses more on the background, while SCGS predominantly focuses on the bird itself. These observations indicate that SCGS can focus more on the instance features, reducing the impact of spurious correlations.

\section{Conclusion}
We propose SCGS, a novel framework that employs clustering and image generation method to reduce the impact of spurious correlations. SCGS can generate high-quality images that meet debiasing requirements. At the same time, SCGS is easy to implement, does not require adding group labels to the data, and can be applied to other debiasing networks without changing the network architecture or algorithms. Experimental results on three biased datasets from different domains demonstrate the effectiveness of SCGS. However, there are still some limitations of our work: when the datasets contain a large number of images, the debiasing effectiveness of SCGS is not significant. Enhancing its debiasing capability on larger datasets is a topic worth exploring.

\bibliographystyle{IEEEtran}
\bibliography{main}
\end{document}